\documentclass{article}
\usepackage{arxiv}

\usepackage[utf8]{inputenc}
\usepackage[T1]{fontenc}
\usepackage{hyperref}
\usepackage{url}
\usepackage{booktabs}
\usepackage{amsfonts}
\usepackage{nicefrac}
\usepackage{microtype}
\usepackage{xcolor}
\usepackage{amsmath}
\usepackage{mathtools}
\usepackage{amsthm}
\usepackage{graphicx}
\usepackage{tikz}
\usepackage{listings}
\usepackage{enumitem}

\theoremstyle{plain}

\theoremstyle{definition}

\theoremstyle{remark}

\makeatletter
\newcommand{\affilstorage}{}
\newcommand{\addtoaffil}[1]{\g@addto@macro\affilstorage{#1}}
\makeatother

\begin{document}

\title{Discovering New Theorems via LLMs with In-Context Proof Learning in Lean}





\author{
  Kazumi Kasaura\\
  OMRON SINIC X Corporation\\
  RIKEN AIP\\
  AutoRes\\
  \texttt{kazumi.kasaura@sinicx.com} \\
  \And
  Naoto Onda\\
  OMRON SINIC X Corporation\\
  NexaScience\\
  AutoRes\\
  \texttt{naoto.onda@sinicx.com} \\
  \And
  Yuta Oriike\\
  CyberAgent\\
  AutoRes\\
  \texttt{oriike\_yuta\_xa@cyberagent.co.jp} \\
  \And
  Masaya Taniguchi\\
  RIKEN AIP\\
  AutoRes\\
  \texttt{masaya.taniguchi@riken.jp} \\
  \And
  Akiyoshi Sannai\\
  Kyoto University\\
  RIKEN AGIS\\
  Shiga University\\
  NII\\
  NISTEP\\
  AutoRes\\
  \texttt{sannai.akiyoshi.7z@kyoto-u.ac.jp} \\
  \And
  Sho Sonoda\\
  RIKEN AIP\\
  CyberAgent\\
  AutoRes\\
  \texttt{sho.sonoda@riken.jp}
}

\date{}

\maketitle


\definecolor{keywordcolor}{rgb}{0.7, 0.1, 0.1}   
\definecolor{tacticcolor}{rgb}{0.0, 0.1, 0.6}    
\definecolor{commentcolor}{rgb}{0.4, 0.4, 0.4}   
\definecolor{symbolcolor}{rgb}{0.0, 0.1, 0.6}    
\definecolor{sortcolor}{rgb}{0.1, 0.5, 0.1}      
\definecolor{attributecolor}{rgb}{0.7, 0.1, 0.1} 

\def\lstlanguagefiles{lstlean.tex}
\lstset{language=lean,
  extendedchars=true,
  frame=single,
}

\begin{abstract}
Large Language Models (LLMs) have demonstrated significant promise in formal theorem proving.
In this study, we investigate the ability of LLMs to discover novel theorems and produce verified proofs. We propose a pipeline called \textit{Conjecturing-Proving Loop} (CPL), which iteratively generates mathematical conjectures and attempts to prove them in Lean 4.
A key feature of CPL is that each iteration conditions the LLM on previously generated theorems and their formal proofs, enabling parameter-free improvement of proof strategies via in-context learning.
We provide both theoretical and experimental evidence that CPL increases the discovery rate of hard-to-prove theorems compared to frameworks that generate statements and proofs simultaneously.
Moreover, our experiments show that reusing the LLM's own formally verified outputs as context consistently improves subsequent proof success, demonstrating the effectiveness of self-generated in-context learning for neural theorem proving.
The source code is available at \url{https://github.com/auto-res/ConjecturingProvingLoop}.
\end{abstract}

\newif\ifCPP
\CPPtrue
\newif\ifDescript
\Descriptfalse
\newif\ifACL
\ACLfalse

\section{Introduction}

Large Language Models (LLMs) have demonstrated significant promise in theorem proving. 
Since LLMs can hallucinate and it is difficult to detect such hallucinations in natural language,
generating formal proofs using an LLM and verifying them using an interactive theorem prover (ITP), such as Lean\footnote{\url{https://lean-lang.org/}}, has been studied.
In this paper, we focus on the ability of LLMs to discover novel theorems.

We propose the \textit{Conjecturing-Proving Loop}, a pipeline for automatically 
generating mathematical conjectures and proving them in Lean 4 format.
By separating the conjecturing and proving phases, we avoid the generation of identical theorems
and encourage proving more difficult theorems.
In other words, CPL employs \emph{stratified sampling} over conjecture/proof candidates to allocate search resources according to proof difficulty, preventing the loop from collapsing to easy, short proofs.
This stratification allows CPL to discover and verify longer proofs that are harder to discover in a simpler framework that samples statements and proofs simultaneously.
In this paper, we present a more detailed theoretical discussion of this point.

Another feature of our approach is that we generate and prove further theorems
using context that includes proven theorems and their proofs,
which enables the generation of more difficult proofs by in-context learning of proof strategies without training of LLMs.
Since the ability of reasoning and generation of Lean code by closed-source LLMs, such as GPT, has been improved recently,
we use them as both conjecturer and prover.
While a disadvantage of using closed-source LLMs is that models cannot be trained freely,
in our framework, the proving ability of LLMs can be improved by in-context learning from previous verified proofs.

In our experiment, when mathematical notions were given as seeds,
we verified whether important properties about them could be rediscovered in our framework.
More specifically, we focused on a few topological notions which are defined only by notions in Mathlib\footnote{\url{https://github.com/leanprover-community/mathlib4}}, Lean's mathematical library, but not included in Mathlib.
We generated theorems about these notions using our framework.
As a result, we found that our framework rediscovered an important theorem about these notions that had been published in mathematical papers, which was not found in the simpler framework without separating the Conjecturer and Prover.
Moreover, we verified that the in-context learning of proof strategies works within our framework:
The important theorem, which cannot be proved without context by LLMs even in natural language, was proved with the generated context.

In summary, the contributions of this paper are as follows. First, we propose the Conjecturing-Proving Loop, a pipeline for automatically generating mathematical conjectures and proving them in Lean 4 format. Second, we demonstrated theoretically and experimentally that our framework enables automatic discovery of hard-to-prove theorems.
Third, we verified that the proving ability of LLMs can be improved by in-context learning, when the verified proofs generated by LLMs themselves before the statement of a target theorem is provided are given as context.

\ifCPP
Our work also suggests the potential for automatic expansion of formal mathematics libraries using AI.
Formalized mathematics is only a part of mathematics expressed in natural language, and expanding formal libraries, such as Mathlib, is crucial for verifying and automating mathematics.
On the other hand, the set of propositions that should be included in the library may not always be obtainable in natural language.
Our framework can generate propositions about given notions while learning them.
\fi
\begin{figure*}[t]
    \centering
    \ifCPP
    \includegraphics[width=\textwidth]{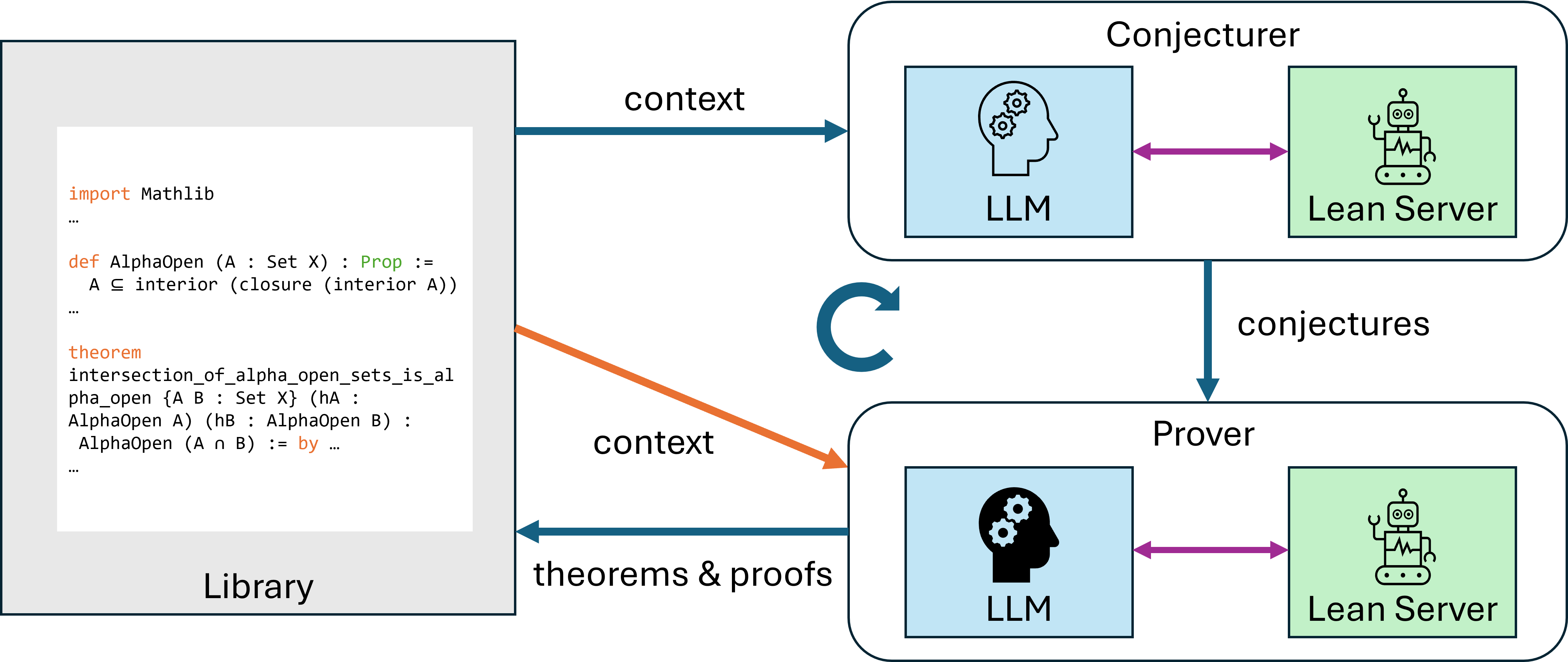}
    \else
    \includegraphics[width=0.8\textwidth]{docs/AIMathematician.png}
    \fi
    \caption{Overview of Conjecturing-Proving Loop. Conjecturer generates conjectures using the library as context, Prover attempts to prove them, and proven conjectures and their proofs are stored in the library as theorems. The library also provides context to Prover.
    Both Conjecturer and Prover processes consist of interactions between LLMs and Lean Server.
    }
    \ifDescript
    \Description{Boxes and arrows}
    \fi
    \label{fig:aifm}
\end{figure*}

\section{Related Work}
There are several works that use LLMs for mathematical reasoning,
both in natural language~\cite{deepseekai2025deepseekv3technicalreport} and formal language for ITP~\cite{ren2025deepseekproverv2advancingformalmathematical,lin2025goedelproverfrontiermodelopensource}.
They focused mainly on solving existing problems and used Supervised Fine-Tuning (SFT) and/or Reinforcement Learning with Verified Rewards (RLVR) to improve problem-solving ability of LLMs.
To overcome the limited dataset for training,
approaches to generate problems to solve by AI are also proposed in some previous
works~\cite{dong2025stpselfplayllmtheorem,huang2025r,zhan2025mathsmith}.
These works are different from our approach in the following two points:
First, our approach is focused on generating and proving meaningful theorems,
while these works are focused on training the LLM prover.
Second, while these works are based on reinforcement learning,
our approach is based on in-context learning, which can be applied to closed-source LLMs.

Several studies have reported that including appropriate contexts in prompts improves the mathematical reasoning ability of LLMs \cite{wei2022chain,zhou2022least,drori2022neural,hu2024minictx,poiroux2025reliable}.
While these studies use handmade examples or data extracted from a database as in-context learning sources, our framework uses outputs from the LLM itself, as in In-Context Reinforcement Learning~\cite{moeini2025survey}.
It has also been proposed to conjecture and prove propositions to be used as lemmas for proving more difficult theorems~\cite{thakur2023context,wang2023lego,chen2025seed,baba2025prover}.
Unlike these studies, we do not provide the theorem to prove to the LLM; instead, we focus on the LLM's ability to discover the theorem and the libraries used as context during proof generation are produced before the target theorem statement is given.

\ifCPP
The technique to leverage feedback from ITP for formal proof generation was proposed~\cite{first2023baldur,thakur2023context,lin2025goedel} and we also adopted it (See \S~\ref{subsection:prover_loop}).
\else
The technique to leverage feedback from ITP for formal proof generation was proposed~\cite{first2023baldur,thakur2023context,lin2025goedel} and we also adopted it (See Appendix~\ref{appendix:prover_loop}).
\fi
However, what we emphasize here is learning strategies from verified proofs of other propositions.

Minimo~\cite{poesia2024learning} shares a similar framework with ours: it jointly trains the conjecturer and prover agents to find theorems. However, while it aims to rediscover mathematics without using existing knowledge, our research has a more practical purpose: attempting to discover theorems by using existing large language models.

A survey~\cite{zhang2026automated} provides a comprehensive and up-to-date summary of theorem generation, including methods that use LLMs.
\section{Method}
In this section, we first present an overview of the framework, and then explain the architectures of the conjecturer and the prover.
\subsection{Pipeline Overview}

Figure~\ref{fig:aifm} illustrates our framework.
Conjecturing-Proving Loop (CPL) consists of four main parts: conjecturer (LLM agent), prover (LLM agent), Lean server, and library (Lean code data).
First, the library is initialized by the user.
\begin{enumerate}
    \item The conjecturer generates novel mathematical conjectures in valid Lean 4 format based on the library, while accessing the Lean server.
    \item For each generated conjecture, the prover tries to generate valid proofs, while accessing the Lean server.
    The library is used as the context also in this step.
    \item The verified pairs of conjectures and their proofs are added to the library. We return to the first step.
\end{enumerate}

\ifCPP
For the details of the conjecturing and proving steps, see the following subsections.
\else
For the details of the second and third steps, see Appendix~\ref{appendix:detail_of_method}.
\fi

By separating the conjecturing and proving phases, we avoid the generation of identical theorems
and encourage proving more difficult theorems.
For a more detailed discussion, please refer to \S~\ref{subsection:discussion}.

The purpose of feeding the library to the conjecturer as the context is
to prevent the generation of duplicate conjectures and
to generate conjectures by analogy from already proven theorems.

The purpose of feeding the library to the prover as the context is
to make already proven theorems available during proof and
to learn proof strategies by in-context learning.

\subsection{Conjecture Loop}
To generate varied conjectures, for each conjecturing step, we use the following process.
\begin{enumerate}[label=\Roman*.]
    \item The conjecturer LLM generates conjectures following the current library.
    \item For each generated conjecture, the Lean server checks whether the conjecture is syntactically valid and novel.
    Verified conjectures are sent to the prover.
\end{enumerate}


The novelty of conjectures is checked using the \texttt{exact?} command in Lean, which checks
whether the conjecture can be proved by existing theorems in the context.
Note that this command is done with context importing the whole Mathlib4 (Lean4 standard library) and including the library and the verified conjectures.
Thus, the checked novelty means that the conjecture is not already in Mathlib4, the already generated library, and the verified conjectures.

The system prompt given to the conjecturer LLM is as follows:

\begin{quote}
You are a contributor to the mathlib4 library. Based on a given library, please generate conjectural new theorems in Lean 4 format; they do not need to be true. Do not generate statements that already appear in the list. Do not include proofs, annotations, or imports. Each new statement should begin with 'theorem' (with no annotations) and end with ':= sorry'. Additionally, use standard mathematical symbols (e.g., $\forall$, $\exists$, $\sqrt{}$) rather than Unicode escape sequences (e.g., \textbackslash u2200).
\end{quote}


\subsection{Prover Loop}\label{subsection:prover_loop}
For each generated conjecture, the prover tries to prove it in the following process.
\begin{enumerate}
    \item The prover LLM produces the proof code of the conjecture. If the LLM judges that the conjecture is not provable, the prover exits the loop as failure.\label{enum:CPL_prover_loop_first_step}
    \item The Lean server verifies the generated proof. If the proof is verified, the prover exits the loop as success.
    \item If the maximum number of trials has been reached, the prover exits the loop with failure.
    Otherwise, the error message from the Lean server is returned to the LLM and we return to step \ref{enum:CPL_prover_loop_first_step}.
\end{enumerate}

The context is given to both the prover and the Lean server.
Thus, not only the prover can learn the proof strategies from the context, but also it can use the theorems in the context as lemmas.

The system prompt given to the prover LLM is as follows:

\begin{quote}
You are a contributor to the mathlib4 library. Please prove the final theorem in the given content using Lean 4. Write the Lean 4 code that directly follows ':=' in the final theorem. The code should either begin with 'by' or be a term expression. You may use the theorems in the given content as lemmas. Do not use 'sorry' in the proof. If you determine that the theorem is not provable, return an empty string instead of a proof. Do not include any additional text.
\end{quote}

In our experiment, the maximum number of trials is set to $16$.

\ifCPP
\subsection{Baseline}
For comparison, we also generated theorems for these notions by a simple loop (SL) framework in which an LLM generates theorems and their proofs at once.
First, the library is initialized by the user.
Unlike CPL, in this simple loop baseline, the conjecturer and the prover are not separated, and the single loop is as follows:
\begin{enumerate}[label=\Roman*]
    \item The LLM generates a statement and its proof in Lean 4 format based on the library, while accessing the Lean server to verify it.\label{enum:SL_loop_first_step}
    \item If the previous step succeeded, the generated pair of statement and proof is stored in the library. We return to step \ref{enum:SL_loop_first_step}.
\end{enumerate}


Step \ref{enum:SL_loop_first_step} is similar to the prover loop.
Its details are as follows:
\begin{enumerate}
    \item The LLM produces a statement and its proof in Lean.\label{enum:first_step_prover_loop}
    \item The Lean server checks the generated content. If it is verified, we exit the loop as success.
    \item If the maximum number of trials has been reached, we exit the loop with failure.
    Otherwise, the error message from the Lean server is returned to the LLM and we return to step \ref{enum:first_step_prover_loop}.
\end{enumerate}

The system prompt given to LLM is as follows:

\begin{quote}
You are a contributor to the mathlib4 library. Based on a given library, please generate a new theorem together with its proof in Lean 4 format. Do not output anything other than the Lean 4 code. The generated code must follow the given library and contain only the theorem statement and its proof. Do not output declarations other than theorem, such as variable, section, or namespace. Do not generate a theorem that already exists in the library. The new theorem should begin with 'theorem' (with no annotations). You may use the theorems in the given library as lemmas in the proof. Do not use 'sorry' in the proof. Additionally, use standard mathematical symbols (e.g., $\forall$, $\exists$, $\sqrt{}$) rather than Unicode escape sequences (e.g., \textbackslash u2200).
\end{quote}

As in the prover loop, the maximum number of trials is set to $16$.

\else
For comparison, we also generated theorems for them by a simple loop framework that LLM generates theorems and its proofs at once,
using chatGPT-o3.
See Appendix~\ref{appendix:simple_loop} for the detail of this baseline.
\fi

\section{Theory}\label{subsection:discussion}
For the following reasons, it is expected that
the distribution of generated theorems differs between SL and CPL.
When a statement and its proof are generated at once, the distribution of generated theorems depends on both the distribution of statements and success rates of proofs.
On the other hand, when multiple proofs are attempted after a statement is generated, the distribution of theorems shifts closer to the distribution of provable statements, and the influence of proof success rates diminishes.

More formally, let $s(T)$ be the probability distribution of statements $T$ generated by the LLM, and let $r(T)$ be the probability that the LLM generates a successful proof of $T$.
We model SL as a simplified process that generates a statement and its proof sequentially, and outputs them if the proof is correct.
CPL is also simplified and modeled as a process that attempts to generate a valid proof, after the generation of a statement, until it succeeds or the number of attempts reaches $N$.
For simplicity, we ignore the influence of contexts.

The probability of generating a theorem $T$ is proportional to $s(T)r(T)$ in SL, and is proportional to $s(T)\left(1-(1-r(T))^N\right)$ in CPL.
Therefore, as $N$ increases, the distribution of theorems in CPL approaches the distribution of provable statements ($T$ such that $r(T)>0$), and even theorems that are difficult to prove tend to be generated more frequently.

On the other hand, while the expected number of proof trials required to discover one theorem in SL is $E_\mathrm{SL}:=\left(\mathbb{E}_{T\sim s}[r(T)]\right)^{-1}$,
it is
\[
E_\mathrm{CPL}:=\frac{\mathbb{E}_{T\sim s}\left[(1-(1-r(T))^N)r(T)^{-1}\right]}{\mathbb{E}_{T\sim s}\left[1-(1-r(T))^N\right]}
\]
in CPL, because, when the statement $T$ is generated, the probability of success in proving $T$ is $1-(1-r(T))^N$ and the expected number of trials is $(1-(1-r(T))^N)r(T)^{-1}$.\footnote{Because $(1-(1-r)^N)r^{-1}$ is actually a polynomial,
we consider its value for $r=0$ as $N$.}

Since $\left(1-(1-r)^N\right)r^{-1}$ is decreasing for $r$, from Chebyshev's sum inequality,
\begin{align*}
 & \mathbb{E}_{T\sim s}\left[(1-(1-r(T))^N)\right]\\
\leq & \mathbb{E}_{T\sim s}\left[(1-(1-r(T))^N)r(T)^{-1}\right]\mathbb{E}_{T\sim s}[r(T)].    
\end{align*}
Thus, $E_{\mathrm{SL}}\leq E_{\mathrm{CPL}}$, which explains why CPL generates fewer theorems than SL.

The condition under which a theorem $T_0$ is more likely to be generated in CPL than in SL under the fixed number of proof trials is as follows.
In SL, at one generation of a statement, the probability to find $T_0$ is $s(T_0)r(T_0)$ and the number of proof trials is always $1$.
In CPL, at one generation of a statement, the probability to find $T_0$ is $s(T_0)(1-(1-r(T_0))^N)$ and the expected number of proof trials is $\mathbb{E}_{T\sim s}\left[(1-(1-r(T))^N)r(T)^{-1}\right]$.
Since $s(T_0)\ll 1$, the desired condition can be approximated as
\[
\frac{1-(1-r(T_0))^N}{\mathbb{E}_{T\sim s}\left[(1-(1-r(T))^N)r(T)^{-1}\right]} > r(T_0),
\]
which is independent of $s(T_0)$.
If $r(T_0)>0$, this can also be written as
\ifACL
\begin{align*}
&(1-(1-r(T_0))^N)r(T_0)^{-1} \\
> & \mathbb{E}_{T\sim s}\left[(1-(1-r(T))^N)r(T)^{-1}\right].    
\end{align*}
\else
\[(1-(1-r(T_0))^N)r(T_0)^{-1}
> \mathbb{E}_{T\sim s}\left[(1-(1-r(T))^N)r(T)^{-1}\right].\]
\fi
Since $\left(1-(1-r)^N\right)r^{-1}$ is decreasing for $r$, provable theorems with sufficiently low proof success rates are more likely to be generated in CPL.
\section{Experiments}
We demonstrated that research-level theorems can be rediscovered by our framework and verified that in-context learning worked effectively in our framework.

\ifACL\else
\ifCPP
\ifDescript\else
The scripts for these experiments and generated libraries are stored at \url{https://github.com/auto-res/ConjecturingProvingLoop}.
\fi\fi
\fi
\subsection{Setting}
In our experiments, we focus on the minor notions in general topology: semi-openness, $\alpha$-openness, and preopenness.
\ifCPP
We used the following file including the definitions of these notions in Lean 4 format as the initial library.
\begin{lstlisting}
import Mathlib
import Aesop

namespace Topology

variable {X : Type*} [TopologicalSpace X]

def P1 (A : Set X) : Prop :=
  A ⊆ closure (interior A)

def P2 (A : Set X) : Prop :=
  A ⊆ interior (closure (interior A))

def P3 (A : Set X) : Prop :=
  A ⊆ interior (closure A)
\end{lstlisting}

\else
We used the definitions of these notions in Lean 4 format (Appendix~\ref{Appendix:seed_file}) as the initial library.
\fi
The notions P1, P2, and P3 are 'semi-open', '$\alpha$-open', and 'preopen', respectively, and are anonymized to prevent LLMs from using existing knowledge.
The reason for choosing these notions is that they can be defined using only notions already present in Mathlib,
while they themselves are not yet included in Mathlib,
and while their mathematical importance has already been recognized and researched,
they are not yet well known enough for LLM to have knowledge of their properties.

We set the following theorem as a target and focused on whether it could be generated or not:
\begin{quote}
\textit{
The intersection of two P2 ($\alpha$-open) sets is P2 ($\alpha$-open)}
\end{quote}
This theorem is important because it is the most difficult part of the proof that $\alpha$-open sets form another topology (Proposition 2 in \cite{njastad1965some}).
We have confirmed that this theorem cannot, at least, be naively derived from the knowledge of the LLM used in the experiment.
See \S~\ref{subsection:prooftarget}.
Whether a generated library contains the desired theorem or not is checked as follows: place the statement of the theorem after the generated library and see if the proof is completed by using \texttt{exact?} command.
If the library contains a proposition that is trivially equivalent to or stronger than the theorem, the completion succeeds.
By doing so, we can accommodate variations in the formulation of the theorem within the range that the Lean server can recognize.

We used GPT-o3\footnote{\url{https://platform.openai.com/docs/models/o3}. While GPT is currently released up to version 5.2, o3 was the latest version when this research began.
Since the notions and theorems used in our experiments were designed assuming o3, GPT-5 was unsuitable because its performance was higher from the start, and thus it was not adopted.}
both in CPL and in SL.
For both CPL and SL, we generated libraries $20$ times as follows: we generated theorems until the API usage reached $14000000$ tokens.

\subsection{Results}
In CPL, 106 theorems were generated on average, and \textbf{the target theorem was discovered 5 times out of 20}. In SL, 328 theorems were generated on average, but \textbf{the target theorem was never generated in any of the 20 runs}.
According to Fisher's exact test ($p=0.024$), CPL is more likely to generate the desired theorem.

An example of generated proofs of this theorem is shown in Appendix~\ref{appendix:proof_of_alpha_open_intersection}. This proof differs from the original proof by Nj\.astad, which suggests that the LLM found this proof independently.


\ifCPP
\else
\begin{figure}[t]
    \begin{minipage}{0.48\textwidth}
        \centering
        \includegraphics[width=\textwidth]{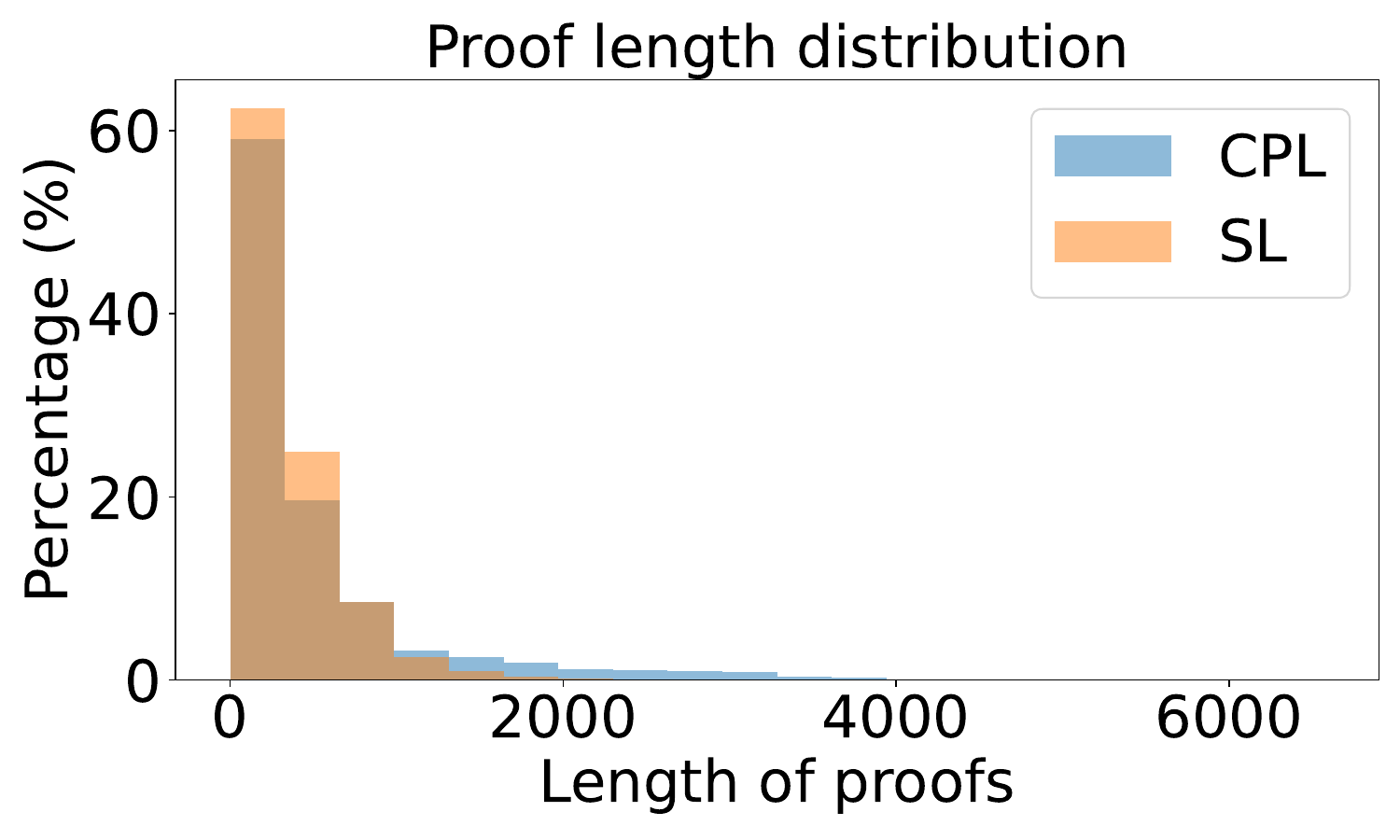}
        \caption{Distribution of proof length of theorems generated by our framework and the simple loop framework, where the size of bins is $10$.}
        \label{fig:proof_lengths}
    \end{minipage}
    \hfill
    \begin{minipage}{0.48\textwidth}
        \centering
        \includegraphics[width=\textwidth]{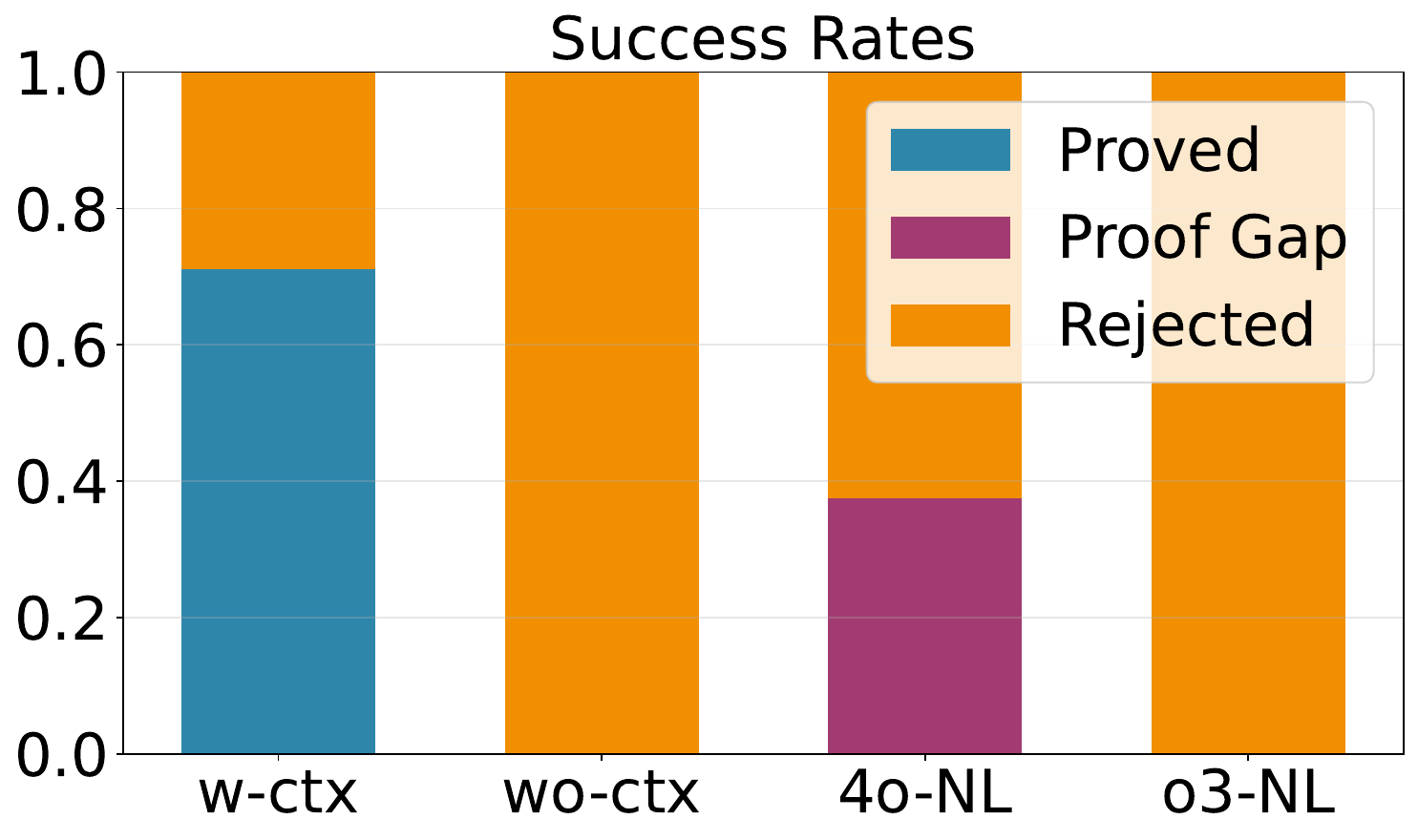}
        \caption{Success rates of proving the theorem by LLMs. With and without context in Lean (\textbf{w-ctx}, \textbf{wo-ctx}), proof in natural language (\textbf{4o-NL}, \textbf{o3-NL}).}
        \label{fig:success_rates}
    \end{minipage}
\end{figure}
\fi

The results showing that while SL generates more theorems, CPL is more likely to generate theorems that are difficult to prove are consistent with the discussion in \S~\ref{subsection:discussion}.
To further verify this, we measured the proof length of the generated theorems.
Figure~\ref{fig:proof_lengths} shows the distribution of proof lengths (the numbers of characters) of the theorems generated by CPL and SL.
It can be observed that CPL can generate theorems with longer proofs than SL.
It is known that there is a positive relationship between the length and difficulty of proofs
\footnote{Note that in this context, “difficulty” refers to the difficulty of generating a valid proof in Lean, and “proof length” refers to the length of the Lean code; these are not necessarily the same as the difficulty or length in natural language.}~\cite{wu2025internlm2,sonoda2026agentic}.
Thus, this result is consistent with the theoretical analysis.

\ifACL
\begin{figure}[t]
    \centering
    \includegraphics[width=1.0\columnwidth]
    {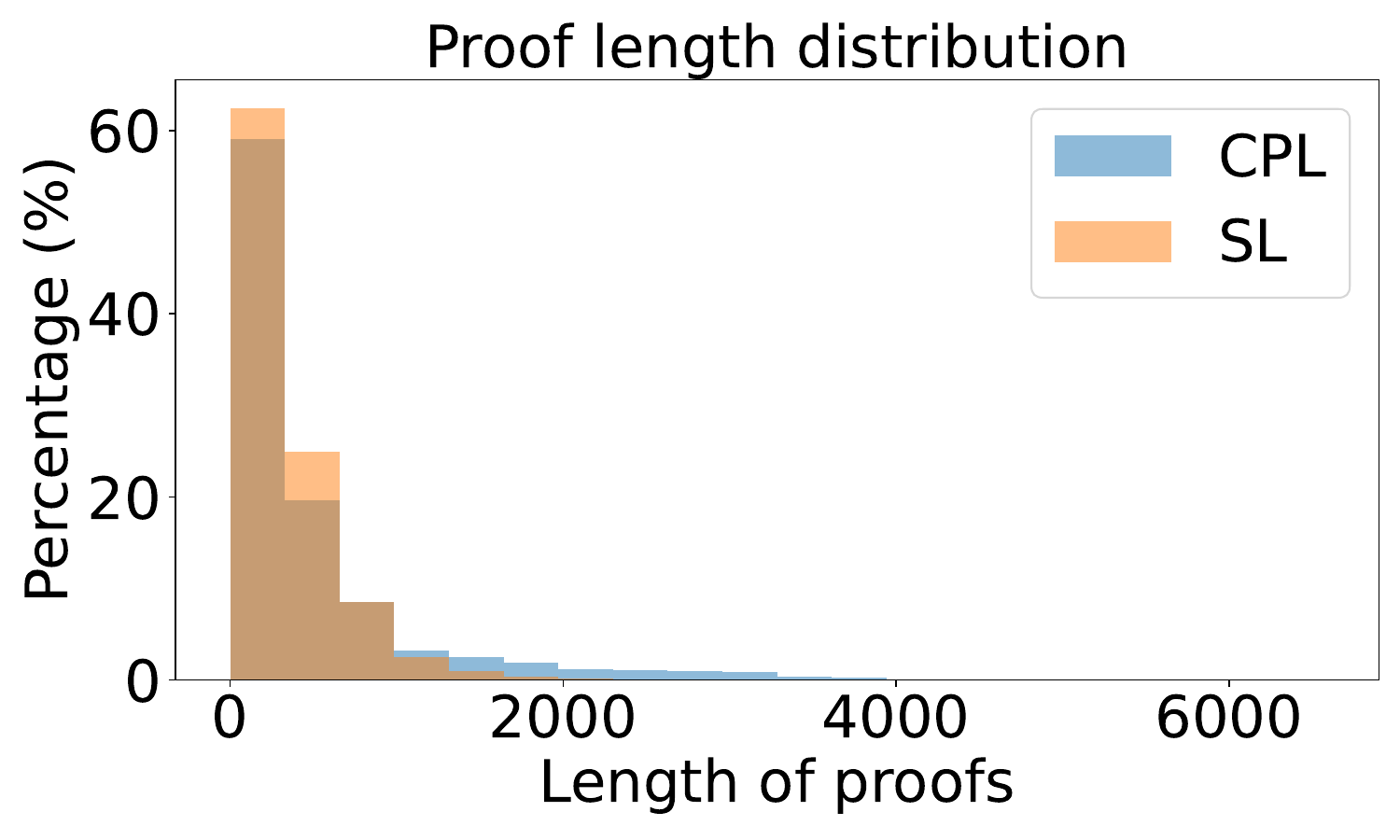}
    \caption{Distribution of proof lengths (the numbers of characters) of theorems generated by our framework and the simple loop framework.}
    \label{fig:proof_lengths}
\end{figure}
\else
\begin{figure}[t]
    \centering
    \includegraphics[width=0.5\columnwidth]
    {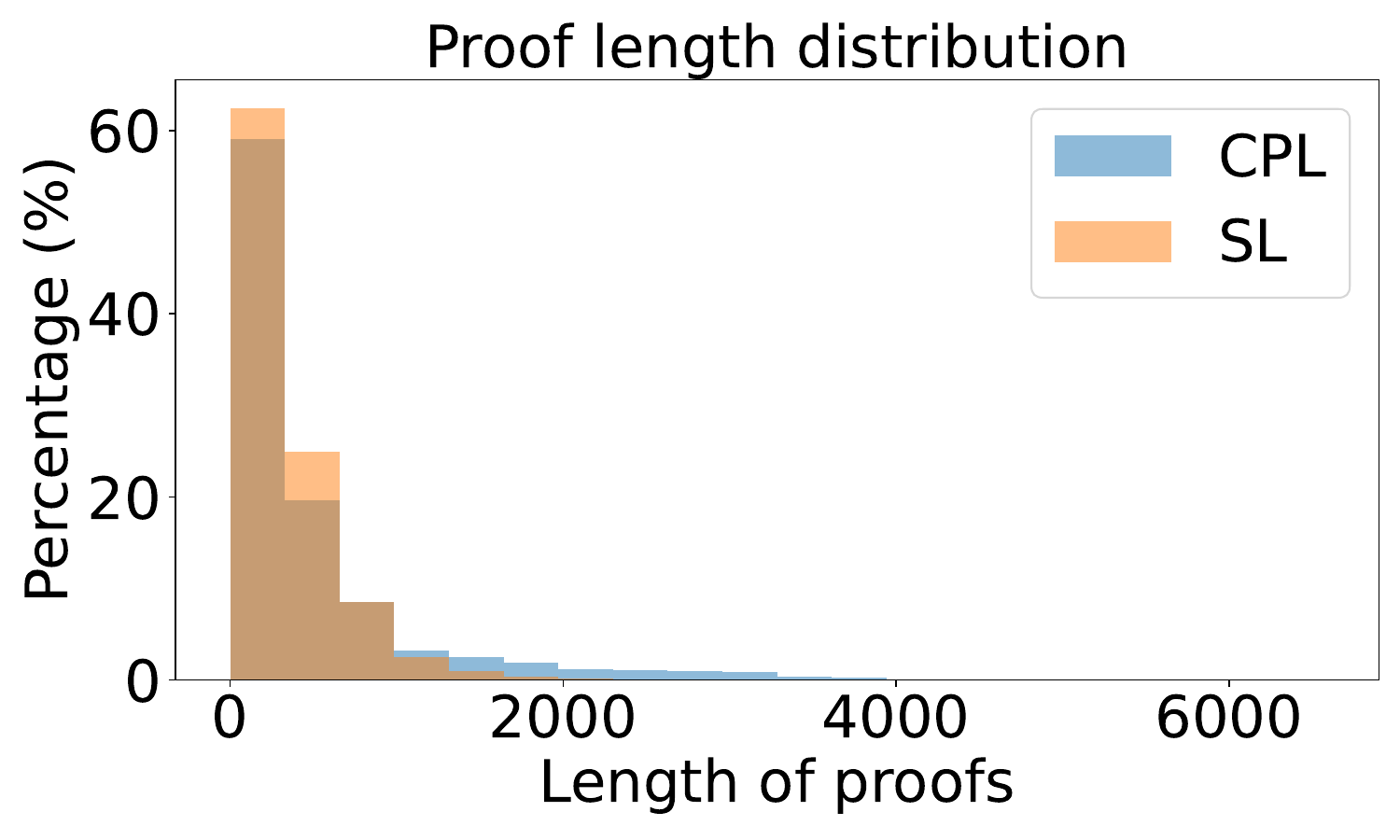}
    \caption{Distribution of proof lengths (the numbers of characters) of theorems generated by our framework and the simple loop framework.}
    \label{fig:proof_lengths}
\end{figure}
\fi
\subsection{Effectiveness of Providing Contexts}
To verify the aforementioned effect of CPL independently, we conducted an additional experiment (\S~\ref{subsection:gen_without_context}).

We also verified that feeding the generated library as a context to the prover improves the proof ability (\S~\ref{subsection:reproving} and \S~\ref{subsection:prooftarget}).

\subsubsection{Generation Without In-Context Learning}\label{subsection:gen_without_context}
To observe the difference between CPL and SL without effects of in-context learning, we generated theorems with only the seed file as a context.
In other words, for both CPL and SL, we independently conducted the first single loops several times, until the API usage reached $3000000$ tokens.

\ifACL
\begin{figure}[t]
    \centering
    \includegraphics[width=1.0\columnwidth]
    {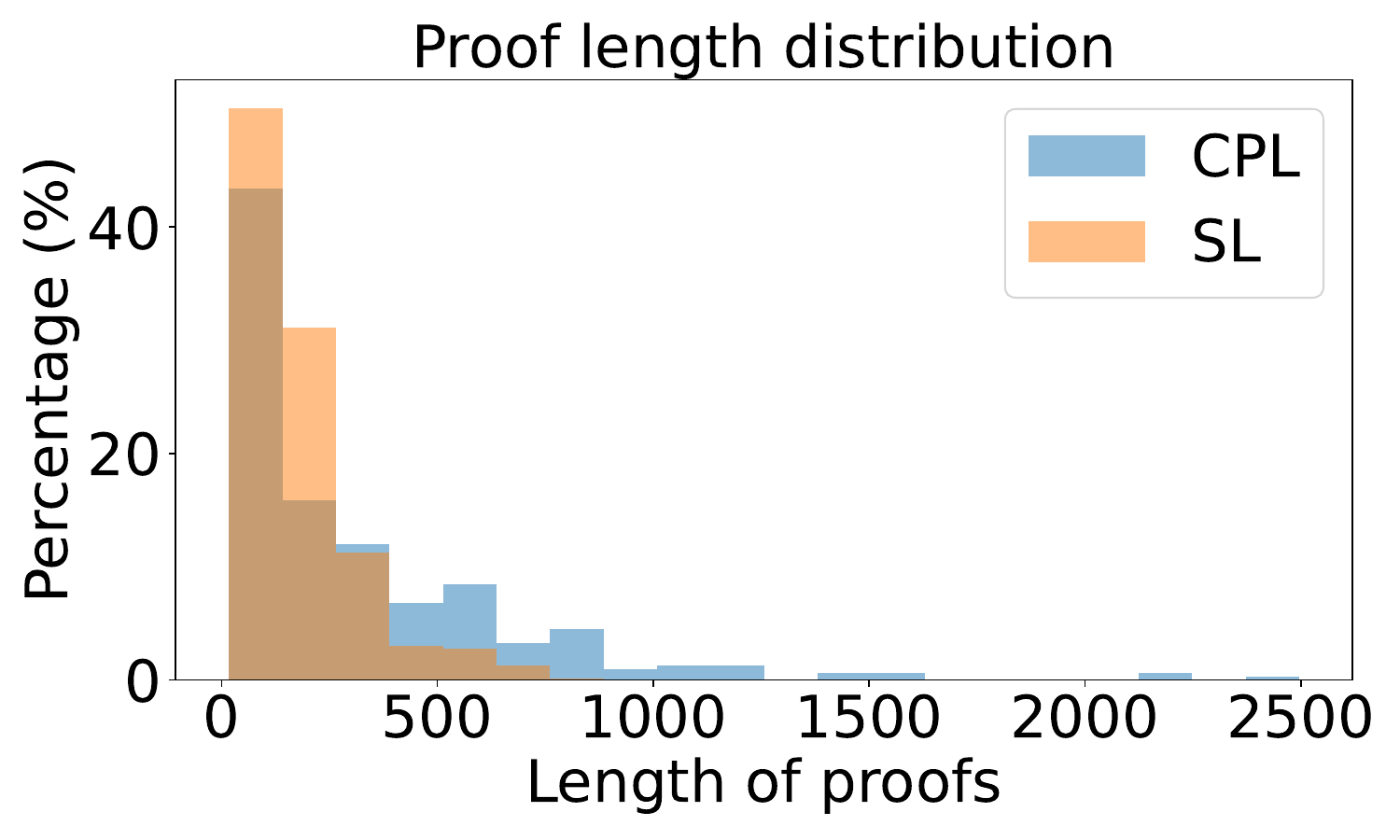}
    \caption{Distribution of proof lengths (the numbers of characters) of theorems generated by our framework and the simple loop framework without in-context learning.}
    \label{fig:proof_lengths_WC}
\end{figure}
\else
\begin{figure}[t]
    \centering
    \includegraphics[width=0.5\columnwidth]
    {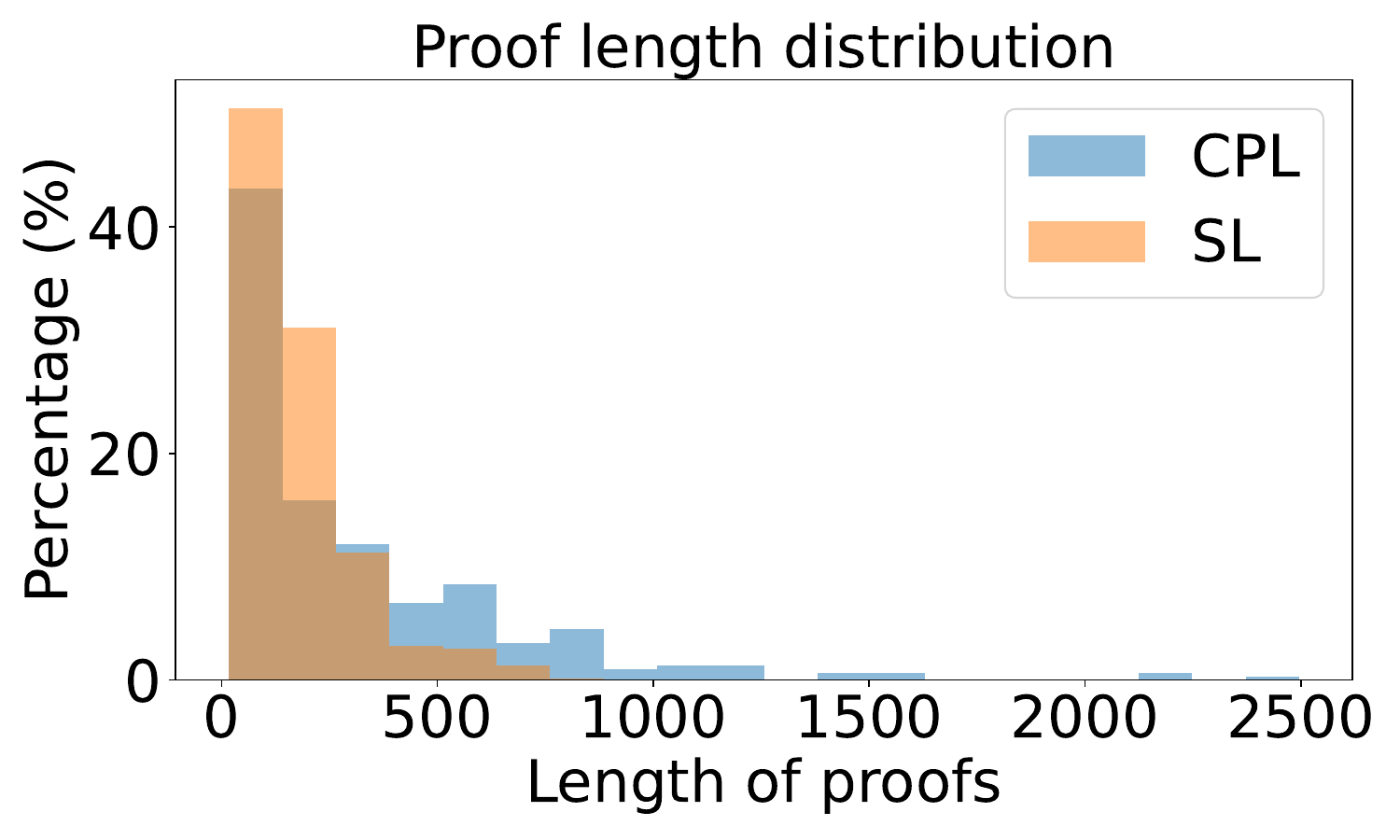}
    \caption{Distribution of proof lengths (the numbers of characters) of theorems generated by our framework and the simple loop framework without context.}
    \label{fig:proof_lengths_WC}
\end{figure}
\fi

Including duplicates, $309$ theorems were generated in CPL and $941$ theorems were generated in SL.
The distributions of generated proofs are shown in Figure~\ref{fig:proof_lengths_WC}.
The shift of distribution is observed.
According to the Kolmogorov–Smirnov test, CPL tends to generate longer proofs than SL with a $p$-value of $1\times 10^{-13}$.

The target theorem was not generated by either CPL or SL. See also the results of the following experiments.

\subsubsection{Reproving Generated Theorems}\label{subsection:reproving}
First, we att\-empted to re-prove all theorems generated in CPL with two settings: one where the context includes the library generated before the theorem to be proved was generated,
and one where the context includes only the definitions of the notions.
As a result, in the setting with contexts, \textbf{99\% of the theorems (2106/2123 theorems)} were proved,
while in the setting without contexts, only \textbf{91\% of the theorems (1935/2123 theorems)} were proved.
According to the McNemar test, this difference is statistically significant at a p-value of $4\times 10^{-35}$.
Thus, the context improves LLMs' proof ability.

\subsubsection{Proof Ability for Alpha-Open Intersection}\label{subsection:prooftarget}
Moreover, we attempted to re-prove the target theorem $16$ times for each of the $5$ contexts in which the target theorem was generated.
(The average number of theorems generated until this theorem was generated is $49$.)
For comparison, we also attempted to re-prove it $80$ times without any generated library.
The procedure is the same as the prover loop, except the system prompt was changed to the following:
\begin{quote}
You are a contributor to the mathlib4 library. Please prove the final theorem in the given content using Lean 4. Write the Lean 4 code that directly follows ':=' in the final theorem. The code should either begin with 'by' or be a term expression. You may use the theorems in the given content as lemmas. Do not use 'sorry' in the proof. If you determine that the theorem is false, return an empty string instead of a proof. Do not include any additional text.
\end{quote}
Note that the condition not to return a proof has changed from "not provable" to "false".

As a result, \textbf{in the setup that included the generated library as context, the re-proof succeeded $7$ times, whereas, in the setup without the library, it failed in all $80$ attempts.}
This suggests that, through in-context learning, the prover acquires the ability to prove theorems that could not be proven without it.

The generated proof of this theorem, shown in Appendix~\ref{appendix:proof_of_alpha_open_intersection}, does not use other generated theorems as lemmas.
Thus, the generated library was used for in-context learning of proof strategies rather than as a collection of lemmas for the proof.

In addition, we also asked LLMs (GPT-4o\footnote{\url{https://platform.openai.com/docs/models/gpt-4o}} and GPT-o3) to prove this theorem in natural language (English)
with context including the definitions of the notions $16$ times and checked the responses by hand.
In the natural language experiment, we used the following system prompt:
\begin{quote}
Please prove the following theorem. If you judge that the theorem is false, please return "False" instead of the proof.
\end{quote}
The statement to prove given to LLMs is as follows:
\begin{quote}
In a topological space, a set is alpha-open if it is a subset of the interior of the closure of its interior. The intersection of any two alpha-open sets is alpha-open.
\end{quote}



As a result, GPT-4o incorrectly stated that the proposition is false $10$ times and generated incorrect proofs $6$ times. GPT-o3 never generated an incorrect proof, but it always judged incorrectly that the theorem is false.
The fact that the majority of judgments in GPT-4o identify the theorem as false implies that the theorem was not included in GPT's knowledge.
An example of a proof with gaps generated by GPT-4o is shown in Appendix~\ref{appendix:incorrect_proof}.

\section{Conclusion and Future Work}
We presented the Conjecturing-Proving Loop, a pipeline for automatically 
generating mathematical conjectures and proving them in Lean 4 format.
We demonstrated that our framework can rediscover a research-level theorem.
We also verified that in-context learning of proof strategies works effectively in our framework.

The propositions highlighted in this study were relatively natural to conjecture.
Future work should focus on refining the conjecture generation process to produce deeper and more insightful mathematical statements, possibly by incorporating techniques for guiding the LLM towards unexplored areas of mathematical theory.

\section*{Acknowledgements}
This work was supported by 
JST Moonshot R\&D Program JPMJMS2236,
JST BOOST JPMJBY24E2, JST CREST JPMJCR2015, %
JSPS KAKENHI 24K21316, 24K16077, %
and
Advanced General Intelligence for Science Program (AGIS), %
the RIKEN TRIP initiative. %

\bibliographystyle{plain}
\bibliography{references}

@misc{ren2025deepseekproverv2advancingformalmathematical,
      title={{DeepSeek-Prover-V2}: Advancing Formal Mathematical Reasoning via Reinforcement Learning for Subgoal Decomposition}, 
      author={Z. Z. Ren and Zhihong Shao and Junxiao Song and Huajian Xin and Haocheng Wang and Wanjia Zhao and Liyue Zhang and Zhe Fu and Qihao Zhu and Dejian Yang and Z. F. Wu and Zhibin Gou and Shirong Ma and Hongxuan Tang and Yuxuan Liu and Wenjun Gao and Daya Guo and Chong Ruan},
      year={2025},
      eprint={2504.21801},
      archivePrefix={arXiv},
      primaryClass={cs.CL},
      url={https://arxiv.org/abs/2504.21801}, 
}

@misc{deepseekai2025deepseekv3technicalreport,
      title={{DeepSeek-V3} Technical Report}, 
      author={{DeepSeek-AI} and Aixin Liu and Bei Feng and Bing Xue and Bingxuan Wang and Bochao Wu and Chengda Lu and Chenggang Zhao and Chengqi Deng and Chenyu Zhang and Chong Ruan and Damai Dai and Daya Guo and Dejian Yang and Deli Chen and Dongjie Ji and Erhang Li and Fangyun Lin and Fucong Dai and Fuli Luo and Guangbo Hao and Guanting Chen and Guowei Li and H. Zhang and Han Bao and Hanwei Xu and Haocheng Wang and Haowei Zhang and Honghui Ding and Huajian Xin and Huazuo Gao and Hui Li and Hui Qu and J. L. Cai and Jian Liang and Jianzhong Guo and Jiaqi Ni and Jiashi Li and Jiawei Wang and Jin Chen and Jingchang Chen and Jingyang Yuan and Junjie Qiu and Junlong Li and Junxiao Song and Kai Dong and Kai Hu and Kaige Gao and Kang Guan and Kexin Huang and Kuai Yu and Lean Wang and Lecong Zhang and Lei Xu and Leyi Xia and Liang Zhao and Litong Wang and Liyue Zhang and Meng Li and Miaojun Wang and Mingchuan Zhang and Minghua Zhang and Minghui Tang and Mingming Li and Ning Tian and Panpan Huang and Peiyi Wang and Peng Zhang and Qiancheng Wang and Qihao Zhu and Qinyu Chen and Qiushi Du and R. J. Chen and R. L. Jin and Ruiqi Ge and Ruisong Zhang and Ruizhe Pan and Runji Wang and Runxin Xu and Ruoyu Zhang and Ruyi Chen and S. S. Li and Shanghao Lu and Shangyan Zhou and Shanhuang Chen and Shaoqing Wu and Shengfeng Ye and Shengfeng Ye and Shirong Ma and Shiyu Wang and Shuang Zhou and Shuiping Yu and Shunfeng Zhou and Shuting Pan and T. Wang and Tao Yun and Tian Pei and Tianyu Sun and W. L. Xiao and Wangding Zeng and Wanjia Zhao and Wei An and Wen Liu and Wenfeng Liang and Wenjun Gao and Wenqin Yu and Wentao Zhang and X. Q. Li and Xiangyue Jin and Xianzu Wang and Xiao Bi and Xiaodong Liu and Xiaohan Wang and Xiaojin Shen and Xiaokang Chen and Xiaokang Zhang and Xiaosha Chen and Xiaotao Nie and Xiaowen Sun and Xiaoxiang Wang and Xin Cheng and Xin Liu and Xin Xie and Xingchao Liu and Xingkai Yu and Xinnan Song and Xinxia Shan and Xinyi Zhou and Xinyu Yang and Xinyuan Li and Xuecheng Su and Xuheng Lin and Y. K. Li and Y. Q. Wang and Y. X. Wei and Y. X. Zhu and Yang Zhang and Yanhong Xu and Yanhong Xu and Yanping Huang and Yao Li and Yao Zhao and Yaofeng Sun and Yaohui Li and Yaohui Wang and Yi Yu and Yi Zheng and Yichao Zhang and Yifan Shi and Yiliang Xiong and Ying He and Ying Tang and Yishi Piao and Yisong Wang and Yixuan Tan and Yiyang Ma and Yiyuan Liu and Yongqiang Guo and Yu Wu and Yuan Ou and Yuchen Zhu and Yuduan Wang and Yue Gong and Yuheng Zou and Yujia He and Yukun Zha and Yunfan Xiong and Yunxian Ma and Yuting Yan and Yuxiang Luo and Yuxiang You and Yuxuan Liu and Yuyang Zhou and Z. F. Wu and Z. Z. Ren and Zehui Ren and Zhangli Sha and Zhe Fu and Zhean Xu and Zhen Huang and Zhen Zhang and Zhenda Xie and Zhengyan Zhang and Zhewen Hao and Zhibin Gou and Zhicheng Ma and Zhigang Yan and Zhihong Shao and Zhipeng Xu and Zhiyu Wu and Zhongyu Zhang and Zhuoshu Li and Zihui Gu and Zijia Zhu and Zijun Liu and Zilin Li and Ziwei Xie and Ziyang Song and Ziyi Gao and Zizheng Pan},
      year={2025},
      eprint={2412.19437},
      archivePrefix={arXiv},
      primaryClass={cs.CL},
      url={https://arxiv.org/abs/2412.19437}, 
}

@misc{dong2025stpselfplayllmtheorem,
      title={{STP}: Self-play {LLM} Theorem Provers with Iterative Conjecturing and Proving}, 
      author={Kefan Dong and Tengyu Ma},
      year={2025},
      eprint={2502.00212},
      archivePrefix={arXiv},
      primaryClass={cs.LG},
      url={https://arxiv.org/abs/2502.00212}, 
}

@inproceedings{lin2025goedelproverfrontiermodelopensource,
title={Goedel-Prover: A Frontier Model for Open-Source Automated Theorem Proving},
author={Yong Lin and Shange Tang and Bohan Lyu and Jiayun Wu and Hongzhou Lin and Kaiyu Yang and Jia LI and Mengzhou Xia and Danqi Chen and Sanjeev Arora and Chi Jin},
booktitle={Second Conference on Language Modeling},
year={2025},
url={https://openreview.net/forum?id=x2y9i2HDjD}
}

@article{zhan2025mathsmith,
  title={MathSmith: Towards Extremely Hard Mathematical Reasoning by Forging Synthetic Problems with a Reinforced Policy},
  author={Zhan, Shaoxiong and Lai, Yanlin and Lu, Ziyu and Lin, Dahua and Yang, Ziqing and Tang, Fei},
  journal={arXiv preprint arXiv:2508.05592},
  year={2025}
}

@article{huang2025r,
  title={R-Zero: Self-Evolving Reasoning {LLM} from Zero Data},
  author={Huang, Chengsong and Yu, Wenhao and Wang, Xiaoyang and Zhang, Hongming and Li, Zongxia and Li, Ruosen and Huang, Jiaxin and Mi, Haitao and Yu, Dong},
  journal={arXiv preprint arXiv:2508.05004},
  year={2025}
}

@article{njastad1965some,
  title={On some classes of nearly open sets},
  author={Nj\.astad, Olav},
  journal={Pacific journal of mathematics},
  volume={15},
  number={3},
  pages={961--970},
  year={1965},
  publisher={Mathematical Sciences Publishers}
}

@article{wei2022chain,
  title={Chain-of-thought prompting elicits reasoning in large language models},
  author={Wei, Jason and Wang, Xuezhi and Schuurmans, Dale and Bosma, Maarten and Xia, Fei and Chi, Ed and Le, Quoc V and Zhou, Denny and others},
  journal={Advances in neural information processing systems},
  volume={35},
  pages={24824--24837},
  year={2022}
}

@article{zhou2022least,
  title={Least-to-most prompting enables complex reasoning in large language models},
  author={Zhou, Denny and Sch{\"a}rli, Nathanael and Hou, Le and Wei, Jason and Scales, Nathan and Wang, Xuezhi and Schuurmans, Dale and Cui, Claire and Bousquet, Olivier and Le, Quoc and others},
  journal={arXiv preprint arXiv:2205.10625},
  year={2022}
}

@article{drori2022neural,
  title={A neural network solves, explains, and generates university math problems by program synthesis and few-shot learning at human level},
  author={Drori, Iddo and Zhang, Sarah and Shuttleworth, Reece and Tang, Leonard and Lu, Albert and Ke, Elizabeth and Liu, Kevin and Chen, Linda and Tran, Sunny and Cheng, Newman and others},
  journal={Proceedings of the National Academy of Sciences},
  volume={119},
  number={32},
  pages={e2123433119},
  year={2022},
  publisher={National Academy of Sciences}
}

@article{moeini2025survey,
  title={A survey of in-context reinforcement learning},
  author={Moeini, Amir and Wang, Jiuqi and Beck, Jacob and Blaser, Ethan and Whiteson, Shimon and Chandra, Rohan and Zhang, Shangtong},
  journal={arXiv preprint arXiv:2502.07978},
  year={2025}
}

@article{thakur2023context,
  title={An in-context learning agent for formal theorem-proving},
  author={Thakur, Amitayush and Tsoukalas, George and Wen, Yeming and Xin, Jimmy and Chaudhuri, Swarat},
  journal={arXiv preprint arXiv:2310.04353},
  year={2023}
}

@inproceedings{first2023baldur,
  title={Baldur: Whole-proof generation and repair with large language models},
  author={First, Emily and Rabe, Markus N and Ringer, Talia and Brun, Yuriy},
  booktitle={Proceedings of the 31st ACM Joint European Software Engineering Conference and Symposium on the Foundations of Software Engineering},
  pages={1229--1241},
  year={2023}
}

@article{lin2025goedel,
  title={Goedel-prover-v2: Scaling formal theorem proving with scaffolded data synthesis and self-correction},
  author={Lin, Yong and Tang, Shange and Lyu, Bohan and Yang, Ziran and Chung, Jui-Hui and Zhao, Haoyu and Jiang, Lai and Geng, Yihan and Ge, Jiawei and Sun, Jingruo and others},
  journal={arXiv preprint arXiv:2508.03613},
  year={2025}
}

@article{poesia2024learning,
  title={Learning formal mathematics from intrinsic motivation},
  author={Poesia, Gabriel and Broman, David and Haber, Nick and Goodman, Noah},
  journal={Advances in Neural Information Processing Systems},
  volume={37},
  pages={43032--43057},
  year={2024}
}

@inproceedings{wu2025internlm2,
  title={InternLM2. 5-stepprover: Advancing automated theorem proving via critic-guided search},
  author={Wu, Zijian and Huang, Suozhi and Zhou, Zhejian and Ying, Huaiyuan and Yuan, Zheng and Zhang, Wenwei and Lin, Dahua and Chen, Kai},
  booktitle={2nd AI for Math Workshop@ ICML 2025},
  year={2025}
}

@article{hu2024minictx,
  title={miniCTX: Neural theorem proving with (long-) contexts},
  author={Hu, Jiewen and Zhu, Thomas and Welleck, Sean},
  journal={arXiv preprint arXiv:2408.03350},
  year={2024}
}

@inproceedings{poiroux2025reliable,
  title={Reliable evaluation and benchmarks for statement autoformalization},
  author={Poiroux, Auguste and Weiss, Gail and Kun{\v{c}}ak, Viktor and Bosselut, Antoine},
  booktitle={Proceedings of the 2025 Conference on Empirical Methods in Natural Language Processing},
  pages={17958--17980},
  year={2025}
}

@article{chen2025seed,
  title={Seed-prover: Deep and broad reasoning for automated theorem proving},
  author={Chen, Luoxin and Gu, Jinming and Huang, Liankai and Huang, Wenhao and Jiang, Zhicheng and Jie, Allan and Jin, Xiaoran and Jin, Xing and Li, Chenggang and Ma, Kaijing and others},
  journal={arXiv preprint arXiv:2507.23726},
  year={2025}
}

@article{baba2025prover,
  title={Prover Agent: An Agent-based Framework for Formal Mathematical Proofs},
  author={Baba, Kaito and Liu, Chaoran and Kurita, Shuhei and Sannai, Akiyoshi},
  journal={arXiv preprint arXiv:2506.19923},
  year={2025}
}

@article{zhang2026automated,
  title={Automated Conjecturing and Theorem Finding: A Survey},
  author={Zhang, Jian and Tan, Si-Cheng},
  journal={Journal of Computer Science and Technology},
  pages={1--21},
  year={2026},
  publisher={Springer}
}

@article{wang2023lego,
  title={Lego-prover: Neural theorem proving with growing libraries},
  author={Wang, Haiming and Xin, Huajian and Zheng, Chuanyang and Li, Lin and Liu, Zhengying and Cao, Qingxing and Huang, Yinya and Xiong, Jing and Shi, Han and Xie, Enze and others},
  journal={arXiv preprint arXiv:2310.00656},
  year={2023}
}

@article{sonoda2026agentic,
  title={Why Agentic Theorem Prover Works: A Statistical Provability Theory of Mathematical Reasoning Models},
  author={Sonoda, Sho and Akiyama, Shunta and Uezato, Yuya},
  journal={arXiv preprint arXiv:2602.10538},
  year={2026}
}

\appendix
\section{Generated Proof of Focused Theorem}\label{appendix:proof_of_alpha_open_intersection}
\lstset{
  frame=none,
}

\begin{lstlisting}
theorem P2_inter {X : Type*} [TopologicalSpace X] {A B : Set X} : P2 A → P2 B → P2 (A ∩ B) := by
  intro hP2A hP2B x hx
  rcases hx with ⟨hxA, hxB⟩
  -- Auxiliary open neighbourhoods furnished by `P2 A` and `P2 B`.
  set UA : Set X := interior (closure (interior A)) with hUA
  set UB : Set X := interior (closure (interior B)) with hUB
  have hUA_open : IsOpen UA := by
    simpa [hUA] using (isOpen_interior :
      IsOpen (interior (closure (interior A))))
  have hUB_open : IsOpen UB := by
    simpa [hUB] using (isOpen_interior :
      IsOpen (interior (closure (interior B))))
  have hxUA : x ∈ UA := by
    have : x ∈ interior (closure (interior A)) := hP2A hxA
    simpa [hUA] using this
  have hxUB : x ∈ UB := by
    have : x ∈ interior (closure (interior B)) := hP2B hxB
    simpa [hUB] using this
  -- Combine the two neighbourhoods.
  have hO_open : IsOpen (UA ∩ UB : Set X) := hUA_open.inter hUB_open
  have hxO : x ∈ UA ∩ UB := ⟨hxUA, hxUB⟩
  -- Main claim: the intersection lies in the relevant closure.
  have hO_sub : (UA ∩ UB : Set X) ⊆ closure (interior (A ∩ B)) := by
    intro y hy
    rcases hy with ⟨hyUA, hyUB⟩
    -- `y` is simultaneously in the closures of `interior A` and `interior B`.
    have hy_clA : y ∈ closure (interior A) := by
      -- `UA ⊆ closure (interior A)`
      have hsub : (UA : Set X) ⊆ closure (interior A) := by
        intro z hz
        have hz' : z ∈ interior (closure (interior A)) := by
          simpa [hUA] using hz
        exact interior_subset hz'
      exact hsub hyUA
    have hy_clB : y ∈ closure (interior B) := by
      have hsub : (UB : Set X) ⊆ closure (interior B) := by
        intro z hz
        have hz' : z ∈ interior (closure (interior B)) := by
          simpa [hUB] using hz
        exact interior_subset hz'
      exact hsub hyUB
    -- Show that every open neighbourhood of `y` meets `interior (A ∩ B)`.
    have : y ∈ closure (interior (A ∩ B)) := by
      apply (mem_closure_iff).2
      intro V hVopen hyV
      -- First refinement, intersect with `UB`.
      have hV1_open : IsOpen (V ∩ UB) := hVopen.inter hUB_open
      have hyV1 : y ∈ V ∩ UB := ⟨hyV, hyUB⟩
      -- Obtain a point in `interior A`.
      rcases (mem_closure_iff).1 hy_clA (V ∩ UB) hV1_open hyV1 with
        ⟨a, ⟨haV, haUB⟩, haIntA⟩
      -- `a` is now in `V`, `UB`, and `interior A`.
      have ha_clB : a ∈ closure (interior B) := by
        have hsub : (UB : Set X) ⊆ closure (interior B) := by
          intro z hz
          have hz' : z ∈ interior (closure (interior B)) := by
            simpa [hUB] using hz
          exact interior_subset hz'
        exact hsub haUB
      -- Second refinement, intersect with `interior A`.
      have hW_open : IsOpen (V ∩ interior A) := hVopen.inter isOpen_interior
      have haW : a ∈ V ∩ interior A := ⟨haV, haIntA⟩
      -- Obtain a point in `interior B`.
      rcases (mem_closure_iff).1 ha_clB (V ∩ interior A) hW_open haW with
        ⟨z, ⟨hzV, hzIntA⟩, hzIntB⟩
      -- `z` lies in `V`, `interior A`, and `interior B`.
      have hzIntAB : z ∈ interior (A ∩ B) := by
        -- The set `interior A ∩ interior B` is open and contained in `A ∩ B`.
        have hS_open : IsOpen (interior A ∩ interior B) :=
          isOpen_interior.inter isOpen_interior
        have hS_sub : (interior A ∩ interior B : Set X) ⊆ A ∩ B := by
          intro w hw
          rcases hw with ⟨h1, h2⟩
          exact ⟨interior_subset h1, interior_subset h2⟩
        have hS_to : (interior A ∩ interior B : Set X) ⊆
            interior (A ∩ B) :=
          interior_maximal hS_sub hS_open
        have hzS : z ∈ interior A ∩ interior B := ⟨hzIntA, hzIntB⟩
        exact hS_to hzS
      exact ⟨z, hzV, hzIntAB⟩
    exact this
  -- Use the neighbourhood just constructed.
  have hNhd : (UA ∩ UB : Set X) ∈ ‽ x :=
    hO_open.mem_nhds hxO
  have h_mem :
      x ∈ interior (closure (interior (A ∩ B))) :=
    (mem_interior_iff_mem_nhds).2
      (Filter.mem_of_superset hNhd hO_sub)
  simpa using h_mem
\end{lstlisting}

\section{Example of Generated Incorrect Proof}\label{appendix:incorrect_proof}
The following is an example of a `proof' of the theorem that the intersection of two alpha-open sets is alpha-open, generated by GPT-4o.
\ifACL
We have only changed the line breaks to adjust the width.
\fi

\begin{quote}
To prove the given theorem, let's first define what it means for a set to be alpha-open in a topological space. 

Let \( (X, \tau) \) be a topological space and let \( A \subseteq X \). The set \( A \) is called alpha-open if 

\[
A \subseteq \text{int}(\overline{\text{int}(A)}),
\]

where \( \text{int}(A) \) denotes the interior of \( A \), and \( \overline{\text{int}(A)} \) denotes the closure of \( \text{int}(A) \).

We aim to prove that the intersection of any two alpha-open sets is alpha-open. Let \( A, B \subseteq X \) be two alpha-open sets; hence, by definition, we have:

1. \( A \subseteq \text{int}(\overline{\text{int}(A)}) \)
2. \( B \subseteq \text{int}(\overline{\text{int}(B)}) \)

We need to show that \( A \cap B \) is also alpha-open, i.e.,

\[
A \cap B \subseteq \text{int}(\overline{\text{int}(A \cap B)}).
\]

Firstly, observe that \( \text{int}(A \cap B) \subseteq \text{int}(A) \cap \text{int}(B) \).

Both closure and interior operations are monotonic, so:

\ifACL
\begin{align*}
\overline{\text{int}(A \cap B)} &\subseteq \overline{\text{int}(A) \cap \text{int}(B)} \\ &\subseteq \overline{\text{int}(A)} \cap \overline{\text{int}(B)}.    
\end{align*}
\else
\[
\overline{\text{int}(A \cap B)} \subseteq \overline{\text{int}(A) \cap \text{int}(B)} \subseteq \overline{\text{int}(A)} \cap \overline{\text{int}(B)}.
\]
\fi

Taking interiors again yields:

\ifACL
\begin{align*}
\text{int}(\overline{\text{int}(A \cap B)}) & \subseteq \text{int}(\overline{\text{int}(A)} \cap \overline{\text{int}(B)}) \\ & \subseteq \text{int}(\overline{\text{int}(A)}) \cap \text{int}(\overline{\text{int}(B)}).
\end{align*}
\else
\[
\text{int}(\overline{\text{int}(A \cap B)}) \subseteq \text{int}(\overline{\text{int}(A)} \cap \overline{\text{int}(B)}) \subseteq \text{int}(\overline{\text{int}(A)}) \cap \text{int}(\overline{\text{int}(B)}).
\]
\fi

By the assumptions that \( A \subseteq \text{int}(\overline{\text{int}(A)}) \) and \( B \subseteq \text{int}(\overline{\text{int}(B)}) \), it follows that

\ifACL
\begin{align*}
A \cap B & \subseteq \text{int}(\overline{\text{int}(A)}) \cap \text{int}(\overline{\text{int}(B)}) \\ & \supset \text{int}(\overline{\text{int}(A \cap B)}).    
\end{align*}
\else
\[
A \cap B \subseteq \text{int}(\overline{\text{int}(A)}) \cap \text{int}(\overline{\text{int}(B)}) \supset \text{int}(\overline{\text{int}(A \cap B)}).
\]
\fi

Therefore,

\[
A \cap B \subseteq \text{int}(\overline{\text{int}(A \cap B)}).
\]

This shows that \( A \cap B \) is indeed alpha-open. Therefore, the intersection of any two alpha-open sets is alpha-open.

Hence, the theorem is true.
\end{quote}

This proof clearly has a gap because, when deriving the last relation from the second-to-last relations, the direction of a subset relation is reversed.

\end{document}